%% file: main.tex
\documentclass[conference]{IEEEtran}
\usepackage[utf8]{inputenc}
\usepackage[T1]{fontenc}
\usepackage{microtype}
\usepackage{graphicx}
\usepackage{booktabs}
\usepackage{tabularx}
\usepackage{amsmath}
\usepackage{cite}
\usepackage{balance}
\usepackage{dblfloatfix}
\usepackage[hidelinks]{hyperref}

\usepackage{xcolor}
\title{Can Loan Processors and Underwriters Spot\\AI-Edited Statements? An Empirical Study}
\author{\IEEEauthorblockN{Simiao Ren, Xingyu (Alex) Shen, Tommy Duong, and Ankit Raj}
\IEEEauthorblockA{Scam.ai (Reality Inc.)\\benren@scam.ai}}
\begin{document}
\maketitle
{\renewcommand{\thefootnote}{}\footnote{Authorship of this paper does not imply employment by or affiliation with Scam.AI (Reality Inc.); some authors are independent researchers who collaborated on this work.}\addtocounter{footnote}{-1}}

\input{sections/short_abstract}
\input{sections/fig_overview}
\input{sections/short_introduction}
\input{sections/fig_examples}
\input{sections/short_study}
\input{sections/short_results}
\input{sections/short_discussion}
\input{sections/short_ethics}
\balance
\bibliographystyle{ieeetr}
\bibliography{references}
\end{document}

%% file: sections/short_abstract.tex
\begin{abstract}
AI-assisted editing can produce loan documents that preserve familiar issuer layouts while changing the applicant's identity. We investigated whether experienced underwriters and loan processors identify such forgeries in first-pass, document-only review. Three practitioners recruited through Upwork each received one applicant package drawn from 134 forged PDFs derived from 47 genuine financial documents supplied with full consent. The edits preserved amounts, dates and balances; identifier inconsistencies introduced by the pipeline were knowingly retained before distribution. Reviewers were not told that files had been altered and returned 132 file-level judgements. Checking their explanations against the originals yielded 14 edit-supported concerns (10.6\%): 12 Suspicious judgements and two Insufficient information judgements. All supported concerns involved identifier disagreements within or across documents, and no reported visual cue identified an actual edit. Thus, approximately one in ten judgements contained evidence of the forgery, despite known inconsistencies remaining in the files. These observations motivate evaluating consistency checks and independent source verification alongside human review. With three reviewer--package pairs and no genuine-file controls, the findings describe this exploratory setting rather than population-wide detection accuracy.
\end{abstract}

%% file: sections/fig_overview.tex
\begin{figure*}[t]
\centering
\includegraphics[width=\textwidth]{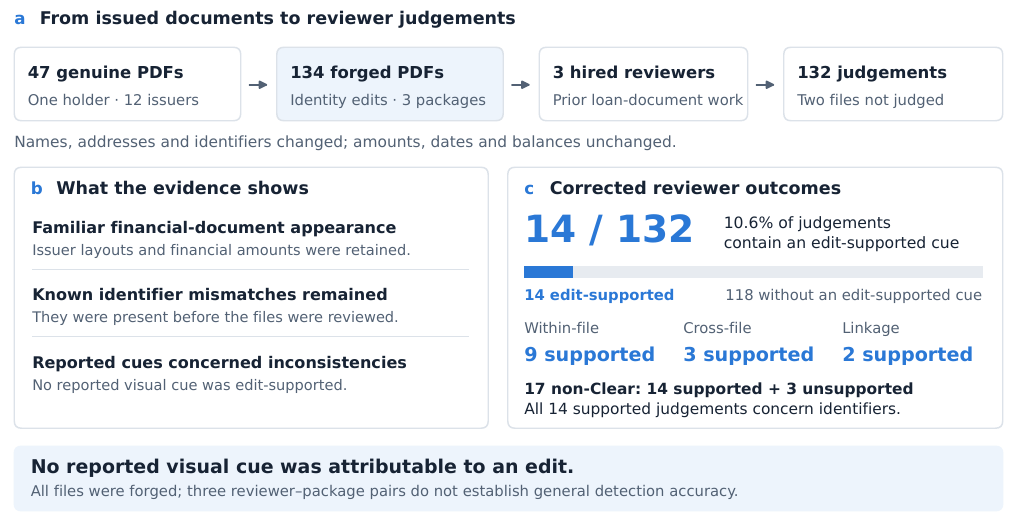}
\caption{\textbf{Study overview: approximately one in ten judgements contained an edit-supported concern.} \textbf{(a)} Agent-driven editing produced 134 forged financial PDFs from 47 genuine source documents; three experienced underwriters or loan processors returned 132 judgements. \textbf{(b)} The files retained familiar issuer layouts and financial amounts, while known identifier inconsistencies were retained before distribution. \textbf{(c)} All 14 supported concerns (10.6\%) involved identifier disagreements: nine within-file, three cross-file and two linkage concerns. No reported visual cue was attributable to an edit.}
\label{fig:pipeline}
\end{figure*}

%% file: sections/short_introduction.tex
\section{Introduction}
A loan applicant's bank statements, pay stubs and tax forms are evidence of identity, income and assets. A reviewer must decide whether the documents are credible enough to use, often before obtaining independent confirmation from their issuers. Financial misrepresentation in lending is well documented~\cite{jiang2014liar,mian2017fraudulent}. AI-assisted PDF editing changes the practical threat: instead of recreating a statement as an image, an agent can call a tool that rewrites selected text while retaining the surrounding native document. Earlier work established that coding agents can perform requested edits to real financial PDFs~\cite{ren2026agents}. The question here is what reviewers notice when those files arrive as an applicant package.

We report a small study designed around that review setting. Forty-seven genuine documents from one consenting holder were identity-swapped into three fictitious applicant packages. Three hired underwriters or loan processors with prior experience each reviewed one package without being told that the documents had been altered. The resulting 132 judgements can be checked against the genuine originals, allowing us to distinguish a concern about the actual forgery from a concern about a property that the original already had.

The central finding is simple (Fig.~\ref{fig:pipeline}): approximately one in ten judgements contained an edit-supported concern. All 14 supported concerns traced to identifier disagreements, such as an account tail that differs between a statement's summary and detail sections, or a taxpayer identifier that differs across forms for the same applicant. No reported concern about appearance was attributable to an edit. Importantly, the pipeline-generated inconsistencies were already known and were knowingly retained in the distributed packages. They supplied opportunities to detect the forgery, rather than being defects discovered only after the study.

This does not show that human review is ineffective in every lending workflow. It shows that, for these reviewers and documents, most identity-swapped files elicited no supported concern, and the successful reports relied on consistency checks rather than a visible editing defect. We focus on this finding and its implications for document-only review, rather than treating a small sample as a benchmark of professional underwriting ability.

\subsection{Related work}
\textbf{Document-image forensics.} DocTamper studies visually consistent text alterations in document images~\cite{qu2023towards}, while receipt-forgery datasets provide annotated financial-document manipulations~\cite{tornes2023receipt}. AIForge-Doc extends evaluation to AI-forged financial and form documents~\cite{wu2026aiforge}, and DOCFORGE-BENCH evaluates pretrained methods across text tampering, receipts and identity documents~\cite{zhao2026docforge}. These works establish relevant detection tasks, but automated image-level performance is a different outcome from the reasons experienced loan-document reviewers report on a native-PDF package.

\textbf{Native PDFs and reasoning beyond appearance.} PDF security research shows how internal document structure can conceal changes~\cite{mainka2021shadow}. PDF-forensics techniques examine evidence beyond the rendered page~\cite{grobler2025technique}; Millet et al. combine internal-object, metadata, overlay and OCR indicators in a multilayer pipeline evaluated on medical certificates~\cite{millet2026multilayer}. The broader identity-document survey by Das et al. distinguishes presentation, injection and generative attacks~\cite{das2026forgeries}. CogniCred studies credential forgeries that require cognitive reasoning and benchmarks multimodal models~\cite{li2026cognicred}. Together, these lines of work motivate distinguishing a convincing appearance from consistency and provenance, without implying that any particular detector would identify our files.

\textbf{Human detection.} Nightingale et al. distinguish detecting a manipulated photograph from locating its alteration~\cite{nightingale2017can}. Groh et al. compare individual, crowd and machine-informed deepfake judgements~\cite{groh2022deepfake}, and Frank et al. establish a cross-country, cross-media human-detection baseline~\cite{frank2024representative}. Their tasks and populations differ from lending, so their rates cannot be transferred to underwriting. Closer document studies examine generated receipts~\cite{zhang2026gpt4oreceipt} and comparisons between image-model financial forgeries and their originals~\cite{wu2026forger}. We instead give experienced reviewers applicant packages without source documents and check whether the concerns they report identify actual edits.

\textbf{Fraud judgement and experience.} Audit research examines experience and explicit fraud-risk assessment in analytical procedures~\cite{knapp2001effects}. Hammersley's review models the role of fraud knowledge and the diagnosticity of risk factors in planning tasks~\cite{hammersley2011review}. This literature motivates studying what evidence reviewers use, rather than equating occupational experience with successful detection. Our observations concern the cues reported by a small experienced sample, not a test of expertise across the profession.

%% file: sections/fig_examples.tex
\begin{figure*}[t]
\centering
\includegraphics[width=\textwidth]{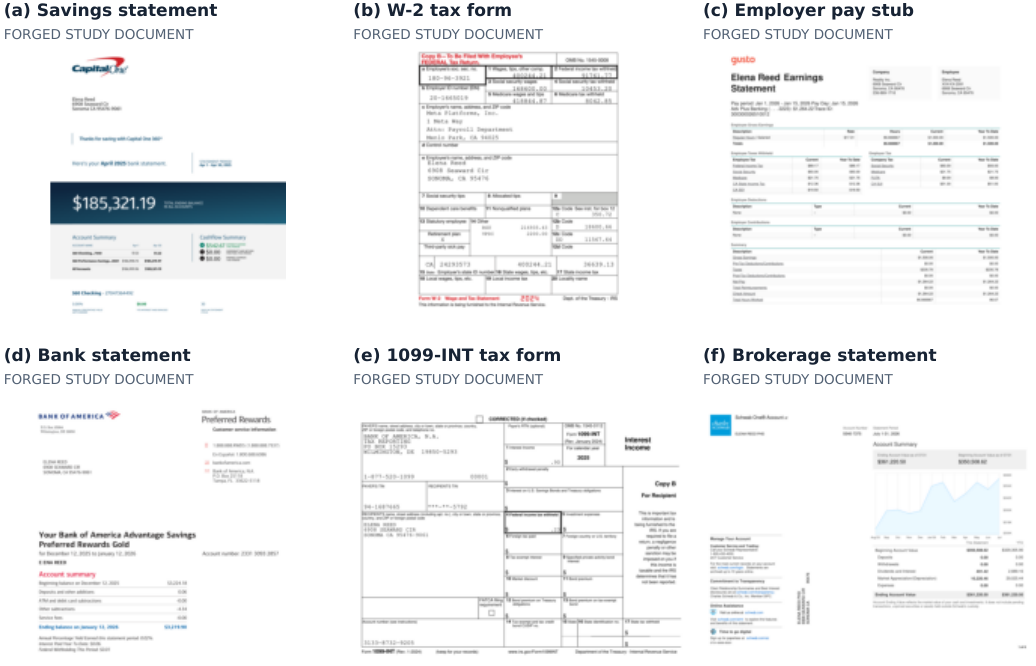}
\caption{Six examples from the forged study set, shown with the author's permission: a savings statement, W-2, employer pay stub, bank statement, 1099-INT and brokerage statement. Names, addresses and identifiers were replaced; financial amounts and issuer layouts were retained from genuine records. These are excerpts of the edited native PDFs, not image-model reconstructions. Known identifier inconsistencies were retained before review.}
\label{fig:examples}
\end{figure*}

%% file: sections/short_study.tex
\section{Study Design}
\label{sec:method}
\subsection{Financial documents and AI-assisted editing}
The source set contains 47 issuer-downloaded PDFs from 12 banks, brokerages, card issuers, employers and payroll providers, issued between 2024 and 2026 to the first author. It includes checking and savings statements, brokerage statements, credit-card statements, balance letters, pay stubs, payslips, W-2s and other tax forms. These document families are relevant to identity, income and asset checks, although the packages also contain material that would not be required for every loan product. The study therefore uses authentic financial layouts and content rather than templates fabricated for an experiment.

AI coding agents used an in-house native-PDF editing tool to replace the holder's identity with three fictitious applicants. The editor retained the issuer's fonts, layout and page geometry. No transaction amount, date or balance was altered. The accepted batch contained 134 PDFs in packages of 47, 42 and 45 files, derived from the 47 genuine sources. These were agent-driven edits to existing PDFs, rather than documents generated as images. Figure~\ref{fig:examples} shows six examples of the resulting financial-document appearance.

\subsection{Known inconsistencies were retained}
The editing pipeline introduced identifier inconsistencies within individual documents and across documents in the same applicant package. The study team knew about these mismatches and retained them before distribution. They were not separately inserted or randomly assigned as experimental conditions. The study therefore evaluates visually preserved financial documents with known opportunities for detection; it does not evaluate packages made fully consistent across every field. The outcome of interest is how often routine reviewers identified the forgery and what kinds of evidence they reported.

\subsection{Reviewers and task}
Three freelancers with prior underwriting or loan-processing experience were recruited through an Upwork posting described as an underwriter research study on documents and statements. The task asked them to follow an ordinary underwriting or loan-processing review, check identity and income information, identify problems including authenticity concerns, and record a judgement and explanation for each file. Reviewers knew their work was for research but were not told that any file had been altered. Each received one applicant package and worked without access to the originals or the other reviewers' results.

All three had previously worked as underwriters or loan processors, as confirmed by the study organiser. Recruitment targeted relevant prior experience with loan-document review. They constitute a small hired sample, not a representative sample of the profession.

Each was paid a fixed fee of US\$25 or US\$30 for a task advertised as taking one to three hours. Actual review time and the use of software aids were not recorded. Two packages were fully reviewed; the third returned judgements on 43 of 45 files, omitting one payslip and one savings statement. These produce 132 returned judgements from 134 supplied PDFs. The sheets used Clear, Suspicious and Insufficient information, with notes about concerns and follow-up.

\subsection{Corrected submission and outcome definition}
One reviewer corrected the assessment file on 21 October 2026 after reporting mislabelled results. Eighteen initial Suspicious labels became Clear, and notes changed on 20 rows; coverage and the two Insufficient information judgements did not change. All results here use that corrected submission. Its native Numbers table and Excel export were checked cell by cell. For another reviewer, the main judgement column is analysed; a separate assessment column calls two Suspicious files Insufficient information. This disagreement does not change whether their stated concerns are attributable to an edit.

We extracted and aligned each forged file's text against its genuine original, then checked every non-Clear judgement's stated reason against those changes. An \emph{edit-supported judgement} contains at least one cue attributable to an edit: an identifier mismatch or a linkage problem introduced by changing the identifier. Appearance and financial-content concerns were checked against the original render. A judgement can therefore be Suspicious without identifying the forgery; conversely, a request for information can contain a supported cue without explicitly calling the file forged.

For a compact description of available cues, we group reviewed files by whether our identifier checks found a within-file inconsistency, only a cross-file inconsistency, or no inconsistency. The checks were grounded in comparison with the genuine originals, so discrepancies attributable to the edits could be distinguished from features already present in the source records. Cue categories were originally coded by the first author and a separate AI-model session, with identical assignments on the 17 non-Clear judgements retained after correction. Only the first author assessed edit support; coding agreement is not independent validation of that assessment.

%% file: sections/short_results.tex
\section{Results}
\label{sec:results}
\subsection{Approximately one in ten judgements was edit-supported}
Table~\ref{tab:overall} gives the pooled result. Of 132 judgements, 14 (10.6\%) contained an edit-supported concern. Twelve supported judgements were Suspicious; the other two were Insufficient information judgements on balance letters whose account could not be linked to the accompanying brokerage statement. Thus, restricting ``caught'' to an explicit Suspicious judgement supported by an edit gives 12/132 (9.1\%). The broader and narrower definitions both place the observed supported rate at approximately 10\%, but they describe different responses.

\begin{table}[t]
\centering
\caption{Pooled outcomes across all three reviewers. A non-Clear judgement need not identify an actual edit.}
\label{tab:overall}
\small
\begin{tabular}{@{}lrr@{}}
\toprule
Returned judgement & Total & Edit-supported \\
\midrule
Clear & 115 & 0 \\
Suspicious & 15 & 12 \\
Insufficient information & 2 & 2 \\
\midrule
All judgements & 132 & 14 (10.6\%) \\
\bottomrule
\end{tabular}
\end{table}

The other 118 judgements contained no edit-supported cue: 115 were Clear, while three were Suspicious for reasons also present in the genuine documents. Counting every Suspicious label as successful detection would therefore overstate how often reviewers identified the actual alterations. Equally, 118 unsupported judgements do not mean that all 118 documents were approved for a loan: follow-up requests and lending decisions were not the study outcome. These counts describe written file-level review responses.

\subsection{All supported concerns involved identifier disagreement}
Nine supported judgements cited within-file account-number disagreements, three cited cross-file taxpayer-identifier disagreements, and two cited an account-linkage failure caused by a changed number. These categories account for all 14 supported judgements. Savings statements supplied repeated within-file account disagreements; cross-file concerns arose on tax forms and balance letters. We pool the observations to describe which cues were reported, rather than compare individual reviewers.

\begin{table}[t]
\centering
\caption{Supported judgements by available inconsistency. Counts refer to reviewed files, not independent editing events.}
\label{tab:tiers}
\small
\begin{tabularx}{\columnwidth}{@{}Xrr@{}}
\toprule
Identifier condition & Judgements & Supported \\
\midrule
Within-file inconsistency & 23 & 9 (39.1\%) \\
Cross-file inconsistency only & 78 & 5 (6.4\%) \\
No inconsistency found & 31 & 0 (0.0\%) \\
\bottomrule
\end{tabularx}
\end{table}

The presence of an inconsistency did not guarantee that it was reported (Table~\ref{tab:tiers}). Within-file inconsistencies yielded 9 supported concerns among 23 judgements. Cross-file inconsistencies yielded 5 among 78: three tax-identifier concerns and two balance-letter linkage concerns. No reviewer reported the account disagreements linking bank statements to payslips. On the 31 reviewed files where our checks found no identifier contradiction, there was no supported concern and no identifier cue was cited at all.

\subsection{Appearance concerns did not identify the edits}
Visual concerns accompanied seven savings-statement judgements supported by account disagreement: transactions looked cropped or edited, and backgrounds looked inconsistent. The genuine statements have the same shaded bands and layout. These seven judgements count as supported because of the account-number discrepancy, not because the visual claim identifies an alteration. One brokerage-statement judgement also questioned an income difference, but that financial feature is present in the original; the supported part of that judgement was its inconsistent account tail.

The three unsupported Suspicious judgements questioned a large transfer from a certificate of deposit, a brokerage statement showing zero balances, and tax-form amounts that appeared to lack leading digits. Comparison with the genuine files confirmed that these were the actual maturity transfer, an empty account and amounts below one dollar. Those concerns did not identify changes made by the editing pipeline. No reviewer cited the original holder's name left in pay-stub bookmarks or the identifying remnants in two filenames. We did not observe whether their PDF viewers displayed the bookmarks.

These findings distinguish a convincing page from a coherent applicant package. The issuer's visual design and genuine financial content survived the identity edit. What reviewers successfully reported was disagreement between pieces of that content, even though the pipeline's output passed its file-level acceptance checks. There was no supported report of a local visual defect in a rewritten name, address or number.

%% file: sections/short_discussion.tex
\section{Discussion}
\subsection{What the result says about loan-document review}
The practical concern is that these were real financial layouts and transaction histories, altered through an agent-driven editing tool and reviewed as applicant evidence. In this setting, roughly 90\% of file judgements contained no cue attributable to the identity edits, despite known identifier mismatches remaining in the packages. The success cases were consistency checks: compare account printings within a statement, compare tax identifiers across forms, or notice that a letter's account no longer links to its statement. The study provides no example of a reviewer identifying the forgery through the appearance of an edited field.

Knowingly retaining the mismatches is central to interpreting the result. The approximately 10\% supported rate was obtained on output that still contained detectable inconsistencies, not on a pipeline engineered to eliminate them. Reusing one coherent replacement for each identifier would remove the specific cues behind the observed supported concerns. It does not follow that reviewers would then detect nothing: they might find other evidence, change their procedures or seek issuer confirmation. A fully consistent package was not tested, so that counterfactual remains open.

\subsection{Implications for verification}
Cross-document reconciliation is a useful response to the cues observed here, but it is not proof of authenticity. Matching names, account tails and taxpayer identifiers can expose mistakes in a forged package; agreement cannot establish that the documents came from the issuer. Legitimate applicants can also have several accounts, changing addresses and different masking formats. A deployed reconciliation system would need to distinguish those cases, and its false-positive rate was not measured in this study.

Commercial document-screening products illustrate the deployment context. Scam.ai describes checks for bank statements, pay stubs and other lending or claims records~\cite{scamai2026documents}; Trustable.ai describes document forensics and evidence-based checks for onboarding and lending~\cite{trustable2026platform}. We cite these websites as descriptions of industry offerings. The present human-review study does not benchmark either platform.

The finding instead motivates a layered workflow: use identifier consistency as a check, and obtain independent evidence when the financial decision requires confidence in provenance. Examples include issuer-confirmed records or account information acquired through a trusted channel. Visual plausibility by itself supplied no supported cue here. We do not infer that a particular commercial verification product would have detected these files, or that every routine lending check was absent; the study recorded reviewer notes, not the full sequence of checks or an actual loan decision.

\subsection{Scope and limitations}
The observations concern three reviewer--package pairs. All three had prior underwriting or loan-processing experience, as confirmed by the study organiser. Each person saw a different applicant package, and two packages omitted source documents because not all planned edits succeeded. We therefore cannot separate reviewer skill from package differences or generalise the 10.6\% rate to the underwriting profession. The outcome is supported concern in written notes, not a controlled measure of recognition under direct questioning.

All files were forged, so the experiment cannot estimate specificity, overall accuracy or the false-positive rate on genuine applicant packages. It also cannot show that 89.4\% of forged files would pass a lender's entire workflow. The edited fields concern identity, rather than amounts or balances; amount manipulation can introduce arithmetic constraints that this experiment did not test. All sources belong to one holder, limiting issuer and applicant diversity.

The available cues also correlate with document family. Most files with no detected identifier inconsistency are credit-card statements, whereas tax forms and payroll records populate the cross-file group. Absence of supported concerns in the former group cannot isolate the effect of consistency from the type of document. Repeated identifiers and repeated reviewer notes further mean that 132 judgements are not 132 independent tests, so we report counts and descriptive percentages rather than inferential comparisons between reviewers.

Finally, actual time on task and use of software were unobserved. Several corrected Clear notes repeat generic payroll wording across different document families; those labels are used as returned, without assuming that the notes establish which checks were performed. Cue support was assessed by the first author, although text differences and the preserved original documents make the decisions auditable. A stronger follow-up would give the same packages to a larger sample of underwriters and loan processors, mix genuine and forged files, compare knowingly inconsistent with consistently edited packages, and record review actions and issuer-verification requests.

\section{Conclusion}
Three hired underwriters or loan processors with prior experience judged 132 native financial PDFs whose identity fields had been altered by an AI-agent-driven editing tool. Only 14 judgements (10.6\%) contained an edit-supported concern, including two requests for information; 12 (9.1\%) were explicitly Suspicious and supported. Every supported concern involved an identifier disagreement that the pipeline generated and the study knowingly retained. No reported visual cue identified an edit. These observations highlight a weakness of relying on page appearance in first-pass financial-document review, and motivate evaluating consistency checks alongside independent verification of document provenance.

%% file: sections/short_ethics.tex
\section*{Ethics and Data Availability}
\label{sec:ethics}
The original financial documents were obtained and used with the full consent of the participants who supplied them. Experienced underwriters and loan processors were recruited through Upwork for relevant prior experience, agreed to research use of their assessments and received the contracted payment. They were not told in advance that the documents had been altered. The original records, full forged PDF files and raw assessment spreadsheets will not be shared. This paper includes selected illustrative excerpts with permission and reports aggregate outcomes. The authors are employees of Scam.ai (Reality Inc.), which develops document-fraud detection products.

Supporting code and the non-restricted resources for this work are available from the Scam.AI Research portal, \url{https://scam.ai/research}~\cite{scamai_research}, with document- and PDF-forgery resources maintained at Trustable, \url{https://trustable.ai}~\cite{trustable_ai}. As noted above, the raw documents used in this study are not publicly released; requests and further details are handled through the Scam.AI Research portal.